\documentclass[11pt,a4paper]{article}
\usepackage[hyperref]{emnlp2018}
\usepackage{times}
\usepackage{latexsym}
\usepackage{amsmath}
\usepackage{graphicx}
    
\usepackage{url}

\aclfinalcopy % Uncomment this line for the final submission
\title{Neural Relation Extraction via Inner-Sentence Noise Reduction and Transfer Learning}

\author{Tianyi Liu$^1$, Xinsong Zhang$^1$, Wanhao Zhou$^1$ and Weijia Jia$^{2,1}$ \\
  $^1$Department of Computer Science and Engineering, Shanghai Jiao Tong University \\
  $^2$Department of Computer and Information Science, University of Macau \\
  {\tt \{liutianyi, xszhang0320, whzhou, jiawj\}@sjtu.edu.cn} \\}

\date{}

\begin{document}
\maketitle
\begin{abstract}
  Extracting relations is critical for knowledge base completion and construction in which distant supervised methods are widely used to extract relational facts automatically with the existing knowledge bases. However, the automatically constructed datasets comprise amounts of low-quality sentences containing noisy words, which is neglected by current distant supervised methods resulting in unacceptable precisions. To mitigate this problem, we propose a novel word-level distant supervised approach for relation extraction. We first build Sub-Tree Parse (STP) to remove noisy words that are irrelevant to relations. Then we construct a neural network inputting the subtree while applying the entity-wise attention to identify the important semantic features of relational words in each instance. To make our model more robust against noisy words, we initialize our network with a priori knowledge learned from the relevant task of entity classification by transfer learning. We conduct extensive experiments using the corpora of New York Times (NYT) and Freebase. Experiments show that our approach is effective and improves the area of Precision/Recall (PR) from 0.35 to 0.39 over the state-of-the-art work.
\end{abstract}

\section{Introduction}
  Relation extraction aims to extract relations between pairs of marked entities in raw texts. Traditional supervised methods are time-consuming for the requirement of large-scale manually labeled data. Thus, \citet{mintz2009distant} propose the distant supervised relation extraction, in which amounts of sentences are crawled from web pages of New York Times (NYT) and labeled with a known knowledge base automatically. The method assumes that if two entities have a relation in a known knowledge base, all instances that mention these two entities will express the same relation. Obviously, this assumption is too strong, since a sentence that mentions the two entities does not necessarily express the relation contained in a known knowledge base. As described in \citet{riedel2010modeling}, the assumption leads to the wrong labeling problem. In order to tackle the wrong labeling problem, various multi-instance learning methods are adopted by mitigating noise between sentences \citep{hoffmann2011knowledge,surdeanu2012multi,zeng2015distant,lin2016neural}. Despite the wrong labeling problem, distant supervised methods may suffer from the low quality of sentences which derive from the large-scale automatically constructed dataset by crawling web pages \citep{yang2017crowdsourced}. To handle the problem of low-quality sentences, we have to face two major challenges: (1) Reduce word-level noise within sentences; (2) Improve the robustness of relation extraction against noise.
  
\begin{figure}[htbp]
    \centering
      \includegraphics[width=7.7cm]{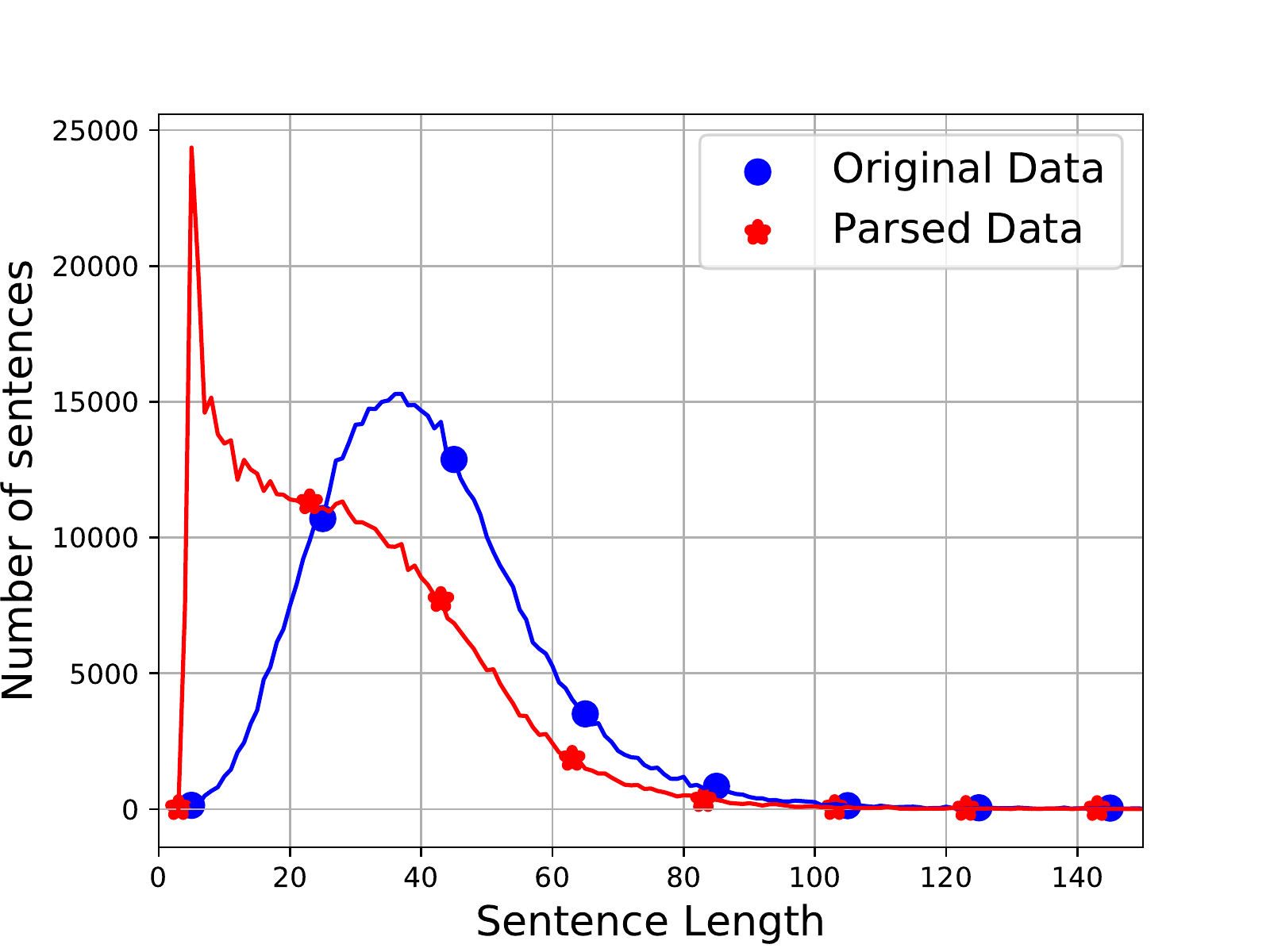}
      \caption{Comparison of sentence length distribution between original data and parsed data.}
      \label{fig:Sent_Dist}
  \end{figure}

  To explain the influence of word-level noise within sentences, we consider the following sentence as an example: \emph{[It is no accident that the main event will feature the junior welterweight champion miguel cotto, a puerto rican, against Paul Malignaggi, an Italian American from Brooklyn.]}, where \emph{Paul Malignaggi} and \emph{Brooklyn} are two corresponding entities. The sub-sentence \emph{[Paul Malignaggi, an Italian American from Brooklyn.]} keeps enough words to express the relation \emph{/people/person/place\_of\_birth}, and the other words could be regarded as noise that may hamper the extractor's performance. Meanwhile, as shown in Figure~\ref{fig:Sent_Dist}, half of the original sentences are longer than 40 words, which means that there are many irrelevant words inside sentences. To be more detail, there are about 12 noisy words in each sentence on average, and 99.4\% of sentences in the NYT-10 dataset have noise. Although the Shortest Dependency Path (SDP) proposed by \citet{xu2015classifying} tries to get rid of irrelevant words for relation extraction, it is not suitable to handle such informal sentences. Moreover, word-level attention has been leveraged to alleviate the impact of noisy words \citep{zhou2016attention}, but it weakens the importance of entity features for relation extraction.

  As for the second challenge, a robust model could extract precise relation features even from low-quality sentences containing noisy words. However, previous neural methods are always lacking in robustness because parameters are initialized randomly and hard to tune with noisy training data, resulting in the poor performance of extractors. Inspired by \citet{kumagai2016learning}, initializing neural networks with a priori knowledge learned from relevant tasks by transfer learning could improve the robustness of the target task. For the relation extraction, entity type classification can be used as the relevant task since entity types provide abundant background knowledge. For instance, the sentence \emph{[Alfead Kahn, the Cornell-University economist who led the fight to deregulate airplanes.]} has a relation \emph{business/person/company}, which is hard to decide without the information that \emph{Alfead Kahn} is a person and \emph{Cornell-University} is a company. Therefore, type features learned from entity type classification are proper a priori knowledge to initialize the relation extractor.

  In this paper, we propose a novel word-level approach for distant supervised relation extraction by reducing inner-sentence noise and improving robustness against noisy words. To reduce inner-sentence noise, we utilize a novel Sub-Tree Parse (STP) method to remove irrelevant words by intercepting a subtree under the parent of entities' lowest common ancestor. As shown in Figure~\ref{fig:Sent_Dist}, the average length of the parsed sentences is much shorter. Furthermore, the entity-wise attention is adopted to alleviate the influence of noisy words in the subtree and emphasize the task-relevant features. To tackle the second challenge, we initialize our model parameters with a priori knowledge learned from the entity type classification task by transfer learning. The experimental results show that our model can achieve satisfactory performance among the state-of-the-art works. Our contributions are summarized as follows:

  \begin{itemize}
    \item To handle the problem of low-quality sentences, we propose the STP to remove noisy words of sentences and the entity-wise attention mechanism to enhance semantic features of relational words.

    \item We first propose to initialize the neural relation extractor with a priori knowledge learned from entity type classification, which strengthens its robustness against low-quality corpus.
    
    \item Our model achieves significant results for distant supervised relation extraction, which improves the Precision/Recall (PR) curve area from 0.35 to 0.39 and increases top 100 predictions by 6.3\% over the state-of-the-art work.
  \end{itemize}
  
  \section{Related Work}
  The distant supervised method plays an increasingly essential role in relation extraction due to its less requirement of human labor \citep{mintz2009distant}. However, an evident drawback of the method is the wrong labeling problem. Thus, multi-instance and multi-label learning methods are proposed to address this issue \citep{riedel2010modeling,hoffmann2011knowledge,surdeanu2012multi}. Meanwhile, other researches \citep{angeli2014combining,han2016global} incorporate human-designed features and leverage Natural Language Processing (NLP) tools.
   
  As neural networks have been widely used, an increasing number of researches have been proposed. \citet{zeng2015distant} use a piecewise convolutional neural network with multi-instance learning. Furthermore, selective attention over instances with the neural network is proposed \citep{lin2016neural}. Making use of entity description, \citet{ji2017distant} assign more precise attention weights. Focused on the imbalance of datasets, a soft label method has been proposed by \citet{liu2017soft}. Recently, reinforcement learning and adversarial learning are widely used to select the valid instances for relation extraction \citep{feng2018reinforcement,P18-1199,P18-1046}.
  
  However, above methods ignore inner-sentence noise. To better remove irrelevant words, the SDP between entities is proved to be effective \citep{de2008stanford,chen2014fast,xu2015classifying,miwa2016end}. Nevertheless, in our observation, the SDP deals with informal texts difficultly (See Section 3.1 for details). Furthermore, word-level attention is adopted to focus on relational words for relation extraction \citep{zhou2016attention}, but it hinders the effect of entity words.
  
  Transfer learning proposed by \citet{pratt1993discriminability} provides a new approach to leverage knoweldge extracted by related tasks to enhance the performance of the target task. Furthermore, parameter transfer learning is proved to be effective to improve the stability of models by initializing model parameters reasonably \citep{pan2010survey,kumagai2016learning}.  

\section{Methodology}
\begin{figure*}[htbp]
    \centering
    \includegraphics[width=16cm]{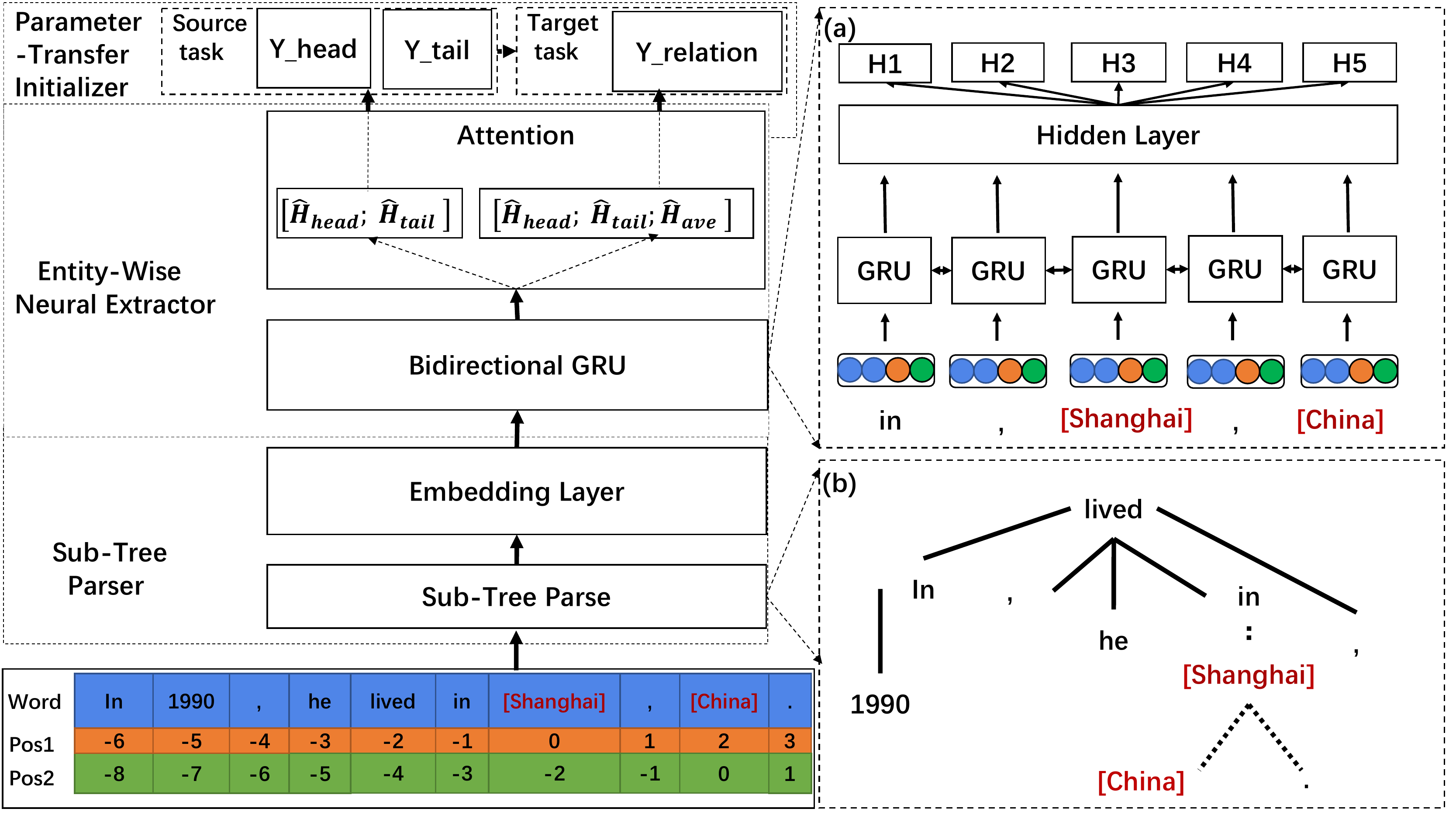}
    \caption{Overall architecture of our model is used for distant supervised relation extraction, expressing the process of handling instances. There are two modules described in detail: (a) One is the BGRU; (b) Another is the STP, where words in the red brackets represent entities (better viewed in color).}
    \label{fig:ARC}
  \end{figure*}

  In this section, we present our methodology for distant supervised relation extraction. Figure~\ref{fig:ARC} shows the overall architecture of our model. Our model is divided into three parts:

  \textbf{Sub-Tree Parser}. Input instances are parsed to dependency parse trees by the Stanford parser\footnote{https://nlp.stanford.edu/software/lex-parser.shtml} \citep{chen2014fast} at first. Then words in the STP and relative positions are transformed to distributed representations.
  
  \textbf{Entity-Wise Neural Extractor}. Given the representation of each subtree, Bidirectional Gated Recurrent Unit (BGRU) extracts specific features. Then, entity-wise attention combined with word-level attention is applied to reducing irrelevant features for relation extraction. Finally, the sentence-level attention is used to alleviate the influence of wrong labeling sentences.

  \textbf{Parameter-Transfer Initializer}. The transfer learning method pre-trains our model parameters from the task of entity type classification aiming at boosting the performance of relation extraction.

  \subsection{Sub-Tree Parser}
  Each instance is put into the dependency parse module to build the dependency parse tree in the first place. Then we can tailor the sentences based on the STP method. Finally, we transform word tokens and position tokens of each instance to distributed representations by embedding matrixes.

  \subsubsection*{Sub-Tree Parse}
  In order to reduce inner-sentence noise and extract relational words, we propose the STP method which intercepts the subtree of each instance under the parent of entities' lowest common ancestor. For instance, in Figure~\ref{fig:ARC}(b), \emph{China} and \emph{Shanghai} are entities connected directly with the appositive relation. The instance \emph{[In 1990, he lives in Shanghai, China.]} will be transformed to \emph{[in Shanghai, China.]} on the basis of the STP, where \emph{in} is the parent of \emph{Shanghai} and \emph{China} lowest common ancestor and kept as important information for expressing the relation \emph{location/location/contain}. Words connected by the imaginary line indicating the extracted subtree are reorganized into their original sequence order to form network inputs.

  Among the parse tree, the SDP has been widely used by \citet{chen2014fast} and \citet{xu2015classifying} to help models focus on relational words. However, in our observation, the SDP is not appropriate in the condition that key relation words are not in the SDP. Although additional information (dependency relations between words) is adopted to enhance the performance of SDP, we found they have the minor effect through our experiment. Thus, we do not make use of other types of linguistic information. As Figure~\ref{fig:ARC}(b) shows, in the SDP method, the original sentence will be transformed to \emph{[Shanghai China]} because \emph{Shanghai} and \emph{China} are connected with each other directly in the dependency parse tree, which results in deleting the keyword \emph{in} and may confuse the model when extracting relations. Compared with SDP, the STP method is more appropriate to extract useful information in informal sentences where relational words are always not in the SDP.
  
  \subsubsection*{Word and Position Embeddings}
  The inputs of the network are word and position tokens, which are transformed to the distributed representations before they are fed into the neural model. We map $j^{th}$ word in the $i^{th}$ instance to a vector of $k$ dimensions denoted as $x_{ij}^w \in R^k$ through the Skip-Gram model \citep{mikolov2013distributed}. Like \citet{zeng2014relation}, we leverage Pos1 and Pos2 to specify entity pairs, which are defined as the relative distances of current word from head entity and tail entity. For instance, in Figure~\ref{fig:ARC} relative distances of \emph{lived} from \emph{Shanghai} and \emph{China} are -2 and -4 respectively. Then, the position token of each word is transformed to a vector in $l$ dimensions. Position embeddings are denoted as $x_{ij}^{p1} \in R^l$ and $x_{ij}^{p2} \in R^l$ respectively. Finally, the input representation for $x_{ij}$ is concatenated by word embedding $x_{ij}^w$, position embeddings $x_{ij}^{p1}$ and $x_{ij}^{p2}$ , which is denoted as $x_{ij}=[x_{ij}^w;x_{ij}^{p1};x_{ij}^{p2}]$ where $x_{ij} \in R^{k+2l}$.

  \subsection{Entity-Wise Neural Extractor}
  As shown in Figure~\ref{fig:ARC}, we transform the STP into feature vectors by BGRU at first. Next, entity-wise attention combined with the hierarchical-level attention mechanism is applied to enhancing semantic features of each instance.

  \subsubsection*{BGRU over STP}
  Since the transfer learning and entity-wise attention require the specific features of entities in tree parsed instances as their input, we adopt Gated Recurrent Unit (GRU) \citep{cho2014learning} to be our based relation extractor, which can extract global information of each word by pointing out its corresponding position in the sequence. It can be briefly described as below:
  \begin{equation}
    h_{it}=GRU(x_{it})
  \end{equation}
  where $x_{it}$ is the $t^{th}$ word representation in the $i^{th}$ parsed instance as described in the input layer, and $h_{it} \in R^m$ is the hidden state of GRU in $m$ dimensions.
  
  Furthermore, BGRU implementing GRU in a different direction can access future as well as past context. Under our circumstance, BGRU combined with the STP can extract semantic and syntactic features adequately. Figure~\ref{fig:ARC}(a) shows the processing of BGRU over STP. The following equation defines the operation mathematically.
  \begin{equation}
    h_{it}=[\overrightarrow{h_{it}} \oplus \overleftarrow{h_{it}}]
  \end{equation}
  In above equation, the $t^{th}$ word output $h_{it} \in R^{m}$ of BGRU is the element-wise addition of the $t^{th}$ hidden states of forward GRU and backward one.

  \subsubsection*{Entity-wise Attention}
  To reduce noise within sentences, we propose the entity-wise attention mechanism to help our model focus on relational words, especially entity words for relation extraction. Assume that $H_i$ is the $i^{th}$ instance matrix consisting of $T$ word vectors $[h_{i1},h_{i2},\cdots,h_{iT}]$ produced by BGRU.

  Not all words contribute equally to the representation of the sentence. Entity words are of great importance because they are significantly beneficial to relation extraction. In our model, entity-wise attention assigns the weight $\alpha_{it}^e$ to focus on the target entity and removes noise further. It is defined as follows:
  \begin{equation}
    \alpha_{it}^e=\begin{cases}
      1 & t=head,tail \\
      0 & others
    \end{cases}  
  \end{equation}
  In the above equation, $\alpha_{it}^e=1$ if $t^{th}$ word belongs to the head or tail entity.

  \subsubsection*{Hierarchical-level Attention}
  To reduce inner-sentence noise further and de-emphasize noisy sentences, we incorporate word-level attention and sentence-level attention as hierarchical-level attention which is introduced in \citet{yang2016hierarchical}.
  
  \textbf{Word-level Attention}. It assigns an additional weight $\alpha_{it}^w$ to relational word $h_{it}$ due to its relevance to the relation as described by \citet{zhou2016attention}. It can be described as follows:
  \begin{equation}
    \alpha_{it}^w=\frac{exp(h_{it}A^wr^w)}{\sum_{t=1}^Texp(h_{it}A^wr^w)}
  \end{equation}
  where $A^w$ is a weighted matrix, and vector $r^w$ can be seen as a high level representation in a fixed query what is the informative word over the other words.
   
  The $i^{th}$ sentence representation $S_i\in R^m$ is computed as a weighted sum of $h_{it}$:
  \begin{equation}
    S_i=\sum_{t=1}^T(\alpha_{it}^w+\alpha_{it}^e)h_{it}  
  \end{equation}

  \textbf{Sentence-level Attention}. After we get the instance representation $S_i$, we adopt the selective attention mechanism over instances to de-emphasize the noisy sentence \citep{lin2016neural}, which is described as follows:
  \begin{align}
    &S=\sum_i\alpha_i^sS_i \label{con:sen_att}\\
    &\alpha_i^s=\frac{exp(S_iA^sr^s)}{\sum_iexp(S_iA^sr^s)}
  \end{align}
  where $A^s$ is a weighted matrix, $r^s$ is the query vector associated with the relation, and $S \in R^m$ is the output of the sentence-level attention layer.

  \subsection{Parameter-Transfer Initializer}
  The transfer learning method pre-trains our model parameters in the entity type classification task, which in turn contributes to the relation extraction.

  \subsubsection*{Pre-learn the Entity Type}
  As entity type information plays a significant role in detecting relation types, the entity type classification task is considered to be the source task, which is learned before the relation extraction task. According to Eq.~\ref{con:sen_att}, outputs of the sentence-level attention layer for the head entity and tail entity task are $S_{head}$ and $S_{tail}$ respectively. They are ultimately fed into the softmax layer:
  \begin{equation}
    \hat{p}^i=softmax(W_iS_i+b_i); \ i\in\{head, tail\}
  \end{equation}
  where $W_i$ and $b_i$ are the weight and bias for the entity type classification task respectively, $\hat{p}^i\in{R^{z_t}}$ is the predicted probability of each class and $z_t$ is the number of entity classes. The loss function of the source task is the negative log-likelihood of the true labels:
  \begin{equation}
    \begin{split}
    &J_e(\theta_0,\theta_{head},\theta_{tail})=\beta\lVert\theta_0\rVert^2\\
    &+\sum_t(-\frac{1}{z_t}\lambda_t\sum_{i=1}^{z_t}y^t_ilog(\hat{p}^t_i)+\beta\lVert\theta_t\rVert^2)\\
    &t\in\{head,tail\}
    \end{split}
  \end{equation}
  where $\lambda_t$ is the weight of each task, $\theta_0$ is the shared model parameters, $\theta_{head}$ and $\theta_{tail}$ are individual parameters for the head and tail entity classification tasks respectively, $y^t\in{R^{z_t}}$ is the one-hot vector representing ground truth, and $\beta$ is the hyper-parameter for $L2$ regularization.
  
  \subsubsection*{Train the Relation Extractor}
  Based on the pre-trained model in the entity type classification task, the relation extractor initializes shared parameters $\theta_0$ within the best state of the pre-trained model and independent parameters $\theta_r$ randomly. Same as the entity type classification task, the output $S_r$ of the attention layer for the relation extraction task is finally fed into the softmax layer and the loss is calculated by cross entropy, which is defined as follows:
  \begin{align}
    &\hat{p}=softmax(W_rS_r+b_r)\\
    \begin{split}
    J_r(\theta_0,\theta_r)=-\frac{1}{z_r}\sum_{i=1}^{z_r}y_ilog(\hat{p}_i)&\\+\beta(\lVert\theta_0\rVert^2+\lVert\theta_r\rVert^2)
    \end{split}
  \end{align}
  where $W_r$, $b_r$, $y\in{R^{z_r}}$, $\hat{p}\in{R^{z_r}}$, $\theta_r$ and $\beta$ are defined similarly in the entity type classification task.

  As shown in Figure~\ref{fig:ARC}, two tasks share all layers except attention and output layers. Assume that the set of total model parameters is $\theta$. Thus, $\theta$, $\theta_0$, $\theta_r$, $\theta_{head}$ and $\theta_{tail}$ have a relationship described in the following equations:
  \begin{align}
    &\theta=\theta_0 \cup \theta_{head} \cup \theta_{tail} \cup \theta_r\\
    \begin{split}
    \theta_i=\{A^w_i,r^w_i,A^s_i,r^s_i,W_i,b_i\} \\
    i \in \{head,tail,r\}
    \end{split}
  \end{align}
  where $A^w_i$, $r^w_i$, $A^s_i$, $r^s_i$, $W_i$ and $b_i$ are parameters in attention and output layers.
  
  \subsubsection*{Optimize the Objective Function}
  At first, we minimize $J$ to obtain $\theta_0$ at the best model state $\hat{\theta}_0$ for entity type classification. Then we minimize $J_r$ for the best performance of relation extraction under the initialization of $\theta_0$ to be $\hat{\theta}_0$. Above process can be summarized as the following equation:
  \begin{equation}
    \begin{split}
    min \quad J(\theta)=&\lambda J_e(\theta_0,\theta_{head},\theta_{tail})+\\
    &(1-\lambda)J_r(\theta_0,\theta_r)
    \end{split}
  \end{equation}
  where $\lambda \in(0,1)$ is the hyperparameter to determine the importance of each task at different training steps. We use Adam \citep{kingma2014adam} optimizer to minimize the objective $J(\theta)$.

\section{Experiments}
  Our experiments are designed to demonstrate that our model alleviates the influence of word-level noise arising from low-quality sentences. In this section, we first introduce the dataset and evaluation metrics. Next, we describe parameter settings. Then we evaluate effects of the STP, entity-wise attention and the parameter-transfer initializer. Finally, we compare our model with the state-of-the-art works by several evaluation metrics.

  \subsection{Dataset and Evaluation Metrics}
  To evaluate the performance of our model, we adopt a widely used dataset NYT-10 developed by \citet{riedel2010modeling}. NYT-10 dataset is constructed by aligning relational facts in Freebase \citep{bollacker2008freebase} with the NYT corpus, where sentences from 2005-2006 are used as training set, and sentences from 2007 are used for testing. For training data, there are 522,611 sentences, 281,270 entity pairs, and 18,252 relational facts; for testing data, there are 172,448 sentences, 96,678 entity pairs and 1,950 relational facts. There are 53 relations including a special relation NA, which means that there is no relation between the entity pair in the instance. Meanwhile, all relations in Freebase are defined on head types and tail types. Therefore, we can construct datasets for type prediction tasks with the same dataset. The dataset has 29 head types and 26 tail types.

  Like previous works, we evaluate our model with the held-out metrics, which compare relations found by models with those in Freebase. The held-out evaluation provides a convenient way to assess models. We report both the PR curve and Precision at top N predictions (P@N) at various numbers of instances under each entity pair:
  
  \textbf{One}: For each entity pair, we randomly select one instance to represent the relation.
  
  \textbf{Two}: For each entity pair, we randomly select two instances to represent the relation.

  \textbf{All}: For each entity pair, we select all instances to represent the relation.

  \subsection{Experimental Settings}
  In the experiment, we utilize word2vec\footnote{https://code.google.com/p/word2vec/} to train word embeddings on NYT corpus. We use the cross-validation to tune our model and grid search to determine model parameters. The grid search approach is used to select optimal learning rate $lr$ for Adam optimizer among $\{0.1,0.001,0.0005,0.0001\}$, GRU size $m \in \{100,160,230,400\}$, position embedding size $l \in \{5,10,15,20\}$. Table~\ref{table:parameter} shows all parameters for our task. We follow experienced settings for other parameters because they make little influence to our model performance.

  \begin{table}[htbp]
    \centering
    \begin{tabular}{|l|c|}
      \hline
      GRU size $m$ & 230\\
      \hline
      Word embedding dimension $k$ & 50\\
      \hline
      POS embedding dimension $l$ & 5\\
      \hline
      Batch size $n$ & 50\\
      \hline
      Entity-Task weights($\lambda_{head},\lambda_{tail}$) & 0.5,0.5\\
      \hline
      Entity-Relation Task weight $\lambda$ & 0.3\\
      \hline
      Learning rate $lr$ & 0.001\\
      \hline
      Dropout probability $p$ & 0.5\\
      \hline
      $l_2$ penalty $\beta$ & 0.0001\\
      \hline
    \end{tabular}
    \caption{Parameter Settings}
    \label{table:parameter}
  \end{table}

  \subsection{Effect of Various Model Parts}
  In this section, we utilize the PR curve to evaluate the effects of three main parts in our model: the STP, entity-wise attention and the parameter-transfer initializer.

  \subsubsection*{Effect of the STP}
  To demonstrate the effect of the STP, we adopt BGRU with Word-Level Attention (WLA) proposed by \citet{zhou2016attention} as our base model. We compare the performance of BGRU, BGRU+STP, and BGRU+SDP.

  \begin{figure}[htbp]
    \centering
    \includegraphics[width=7.7cm]{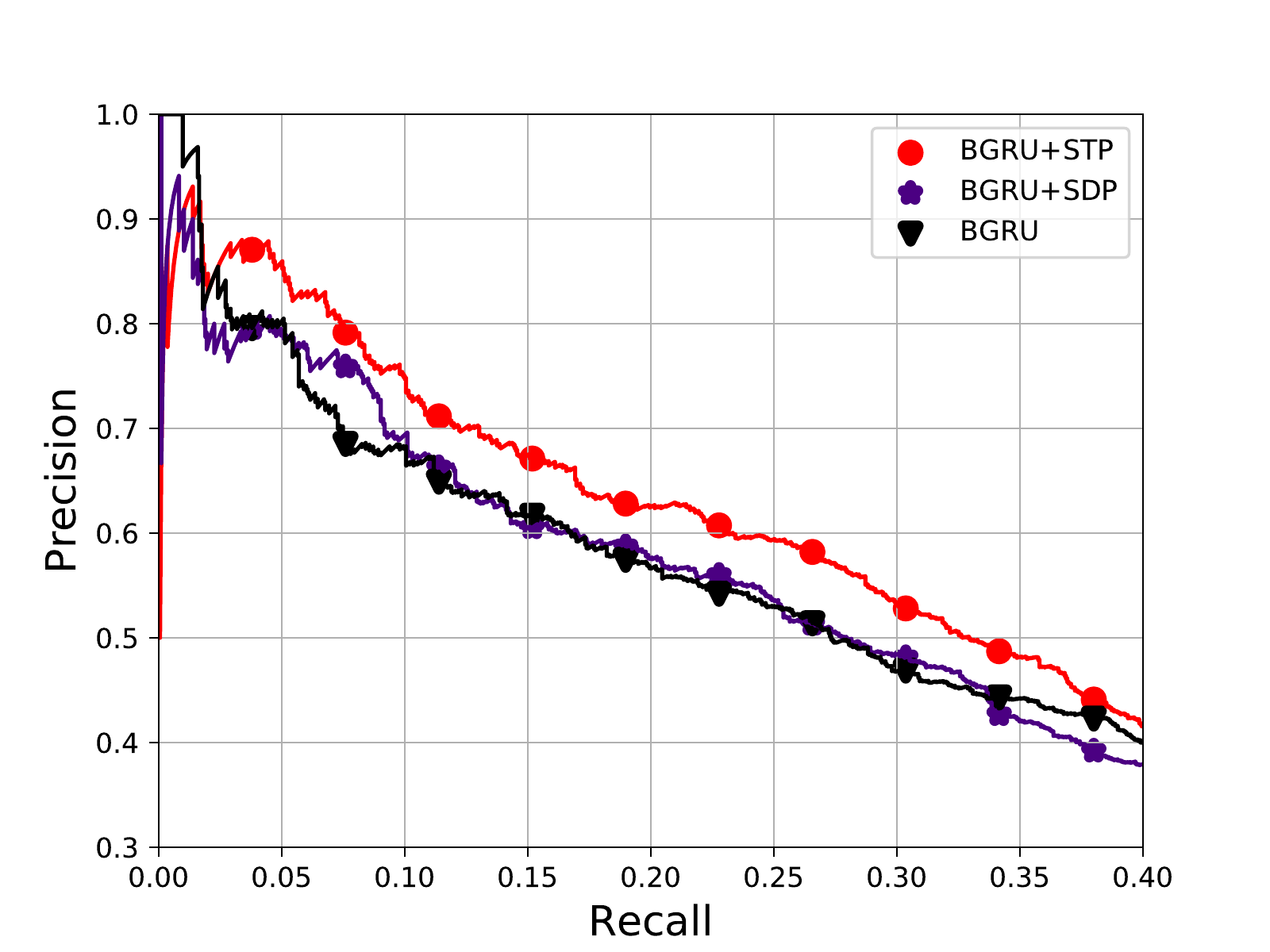}
    \caption{PR curves for BGRU, BGRU+SDP and BGRU+STP.}
    \label{fig:new_or_old}
  \end{figure}

  From Figure~\ref{fig:new_or_old}, we can observe that the model with the STP performs best, and the SDP model obtains an even worse result than the pure one. The PR curve areas of BGRU+SDP and BGRU are about $0.332$ and $0.337$ respectively, while BGRU+STP increases it to $0.366$. The result indicates: (1) Our STP can get rid of irrelevant words in each instance and obtain more precise sentence representation for relation extraction. It proves that our STP module is effective. (2) The SDP method is not appropriate to handle low-quality sentences where key relation words are not in the SDP.

  \subsubsection*{Effect of Entity-wise Attention}

  \begin{table}[htbp]
    \centering
    \begin{tabular}{|c|c|c|}
      \hline
      Test Settings & \multicolumn{2}{|c|}{PR Curve Area} \\
      \hline
      Dataset & Original Data & STP Data \\
      \hline
      BGRU & 0.337 & 0.366 \\
      \hline
      -WLA+EWA & 0.365 & 0.375 \\
      \hline
      +EWA & \textbf{0.372} & \textbf{0.383} \\
      \hline
    \end{tabular}
    \caption{PR curve areas for BGRU, BGRU-WLA+EWA and BGRU+EWA on various datasets.}
    \label{table:attention}
  \end{table}

  To evaluate the effect of entity-wise attention combined with word-level attention, we utilize BGRU in three settings on our tree parsed data and original data. One setting is to use WLA mechanism only (BGRU). The second one is to replace WLA with the Entity-Wise Attention (EWA) mechanism (BGRU-WLA+EWA). The third one is to incorporate two mechanisms (BGRU+EWA).

  \begin{figure}[htbp]
    \centering
    \includegraphics[width=7.7cm]{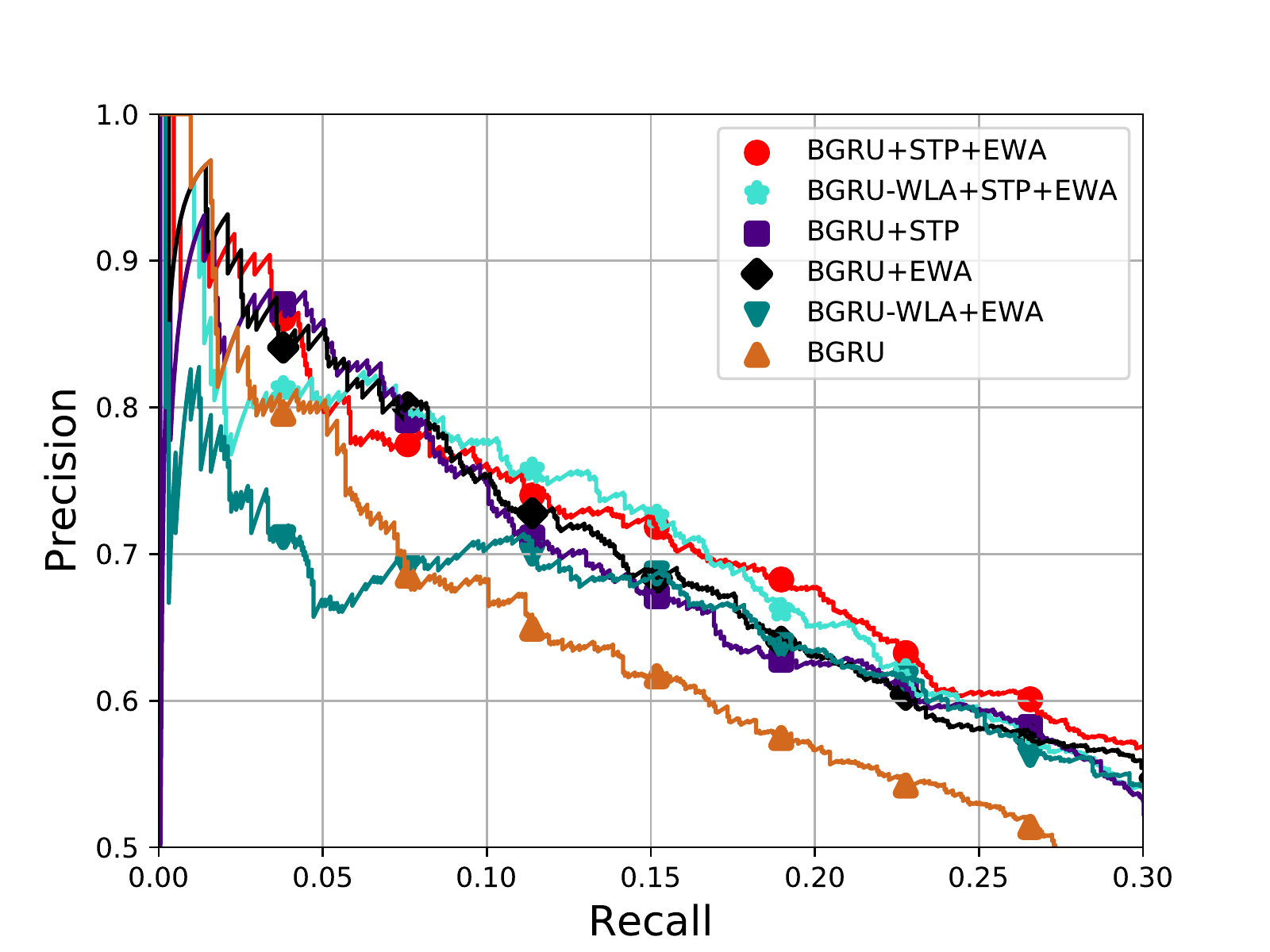}
    \caption{PR curves for BGRU, BGRU-WLA+EWA and BGRU+EWA on various datasets.}
    \label{fig:attention}
  \end{figure}

  \begin{table*}[htbp]
    \centering
    \resizebox{\textwidth}{!}{
    \begin{tabular}{|c|c|c|c|c|c|c|c|c|c|c|c|c|}
      \hline
      Test Settings & \multicolumn{4}{c}{One} & \multicolumn{4}{|c|}{Two}  & \multicolumn{4}{c|}{All} \\
      \hline
      P@N & 100 & 200 & 300 & Mean & 100 & 200 & 300 & Mean & 100 & 200 & 300 & Mean \\
      \hline
      Mintz & 35.0 & 37.5 & 37.3 & 36.6 & 51.0 & 42.0 & 43.3 & 45.4 & 54.0 & 50.5 & 45.3 & 49.9\\
      \hline
      MultiR & 64.0 & 61.5 & 53.7 & 59.7 & 62.0 & 61.5 & 58.7 & 61.1 & 75.0 & 65.0 & 62.0 & 67.3\\
      \hline
      MIML & 62.0 & 59.0 & 54.7 & 58.6 & 69.0 & 59.5 & 59.0 & 62.5 & 70.0 & 64.5 & 60.3 & 64.9\\
      \hline
      PCNN  & 73.3 & 64.8 & 56.8 & 65.0 & 70.3 & 67.2 & 63.1 & 66.9 & 72.3 & 69.7 & 64.1 & 68.7 \\
      \hline
      PCNN+ATT & 78.0 & 68.0 & 60.7 & 68.9 & 75.0 & 74.0 & 66.3 & 71.8 & 82.0 & 74.0 & 69.0 & 75.0 \\
      \hline
      BGRU & 72.0 & 62.5 & 59.0 & 64.5 & 70.0 & 64.0 & 64.7 & 66.2 & 74.0 & 68.0 & 65.0 & 69.0 \\
      \hline
      +STP & 73.0 & 63.0 & 60.7 & 65.6 & 83.0 & 72.5 & 68.0 & 74.5 & 86.0 & 76.0 & 70.3 & 77.4 \\
      \hline
      +EWA & 82.0 & 71.5 & 66.3 & 73.3 & 84.0 & 79.5 & 70.3 & 77.9 & 86.0 & 81.5 & 75.3 & 80.9 \\
      \hline
      +TL & \textbf{83.0} & \textbf{75.5} & \textbf{67.0} & \textbf{75.2} & \textbf{85.0} & \textbf{81.0} & \textbf{72.3} & \textbf{79.4} & \textbf{87.0} & \textbf{83.0} & \textbf{78.0} & \textbf{82.7} \\
      \hline
    \end{tabular}}
    \caption{P@N for relation extraction in the entity pairs with different number of sentences}
    \label{table:P@N}
  \end{table*}

  From Table~\ref{table:attention} and Figure~\ref{fig:attention}, we can obtain: (1) Regardless of the dataset that we employ, BGRU-WLA(+STP)+EWA outperforms BGRU(+STP). To be more specific, the PR curve area has a relative improvement of over $2.3\%$, which demonstrates that entity-wise hidden states in the BGRU present more precise relational features than other word states. (2) BGRU(+STP)+EWA achieves further improvements and outperforms the baseline by over $4.6\%$, because it considers more information than entity or relational words alone. Thus, it indicates that entity words are essential for relation extraction, but they can not represent features of the whole sentence without other words.

  \subsubsection*{Effect of Parameter-Transfer Initializer}

  \begin{figure}[htbp]
    \centering
    \includegraphics[width=7.7cm]{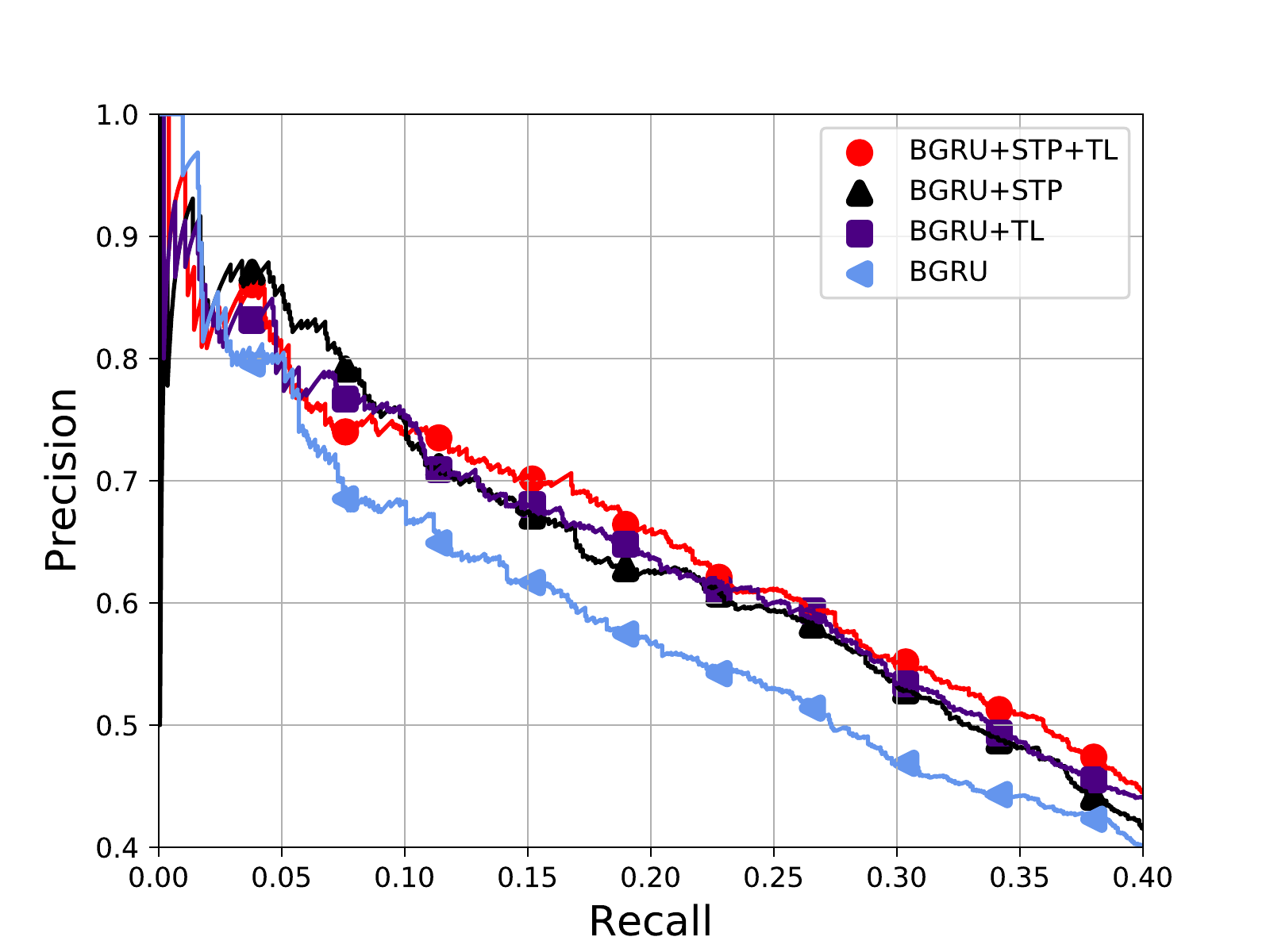}
    \caption{PR curves for BGRU, BGRU+TL, BGRU+STP and BGRU+STP+TL}
    \label{fig:tl}
  \end{figure}

  To evaluate the effect of the parameter-transfer initializer in our model, we leverage BGRU under four circumstances. The first one is to directly apply it on the original dataset. The second one tests BGRU combined with Transfer Learning (TL) on the original dataset. The third one uses BGRU on our STP dataset. The fourth one examines BGRU+TL on our STP dataset.

  From Figure~\ref{fig:tl}, we can conclude: (1) Regardless of the dataset that we use, models with TL achieve better performance, which improve the PR curve area by over $4.7\%$. It demonstrates that transfer learning helps our model become more robust against noise. (2) BGRU+STP+TL achieves the best performance and increases the area to $0.383$, while areas of BGRU, BGRU+STP and BGRU+TL are 0.337, 0.366 and 0.372 respectively. It means that the TL method works well with the STP and can resist noisy words further. 
  
  \subsection{Comparison with Baselines}
  To evaluate our approach, we select the following six methods as our baseline:
  
  \textbf{Mintz} \citep{mintz2009distant} proposes the human-designed feature model.
  
  \textbf{MultiR} \citep{hoffmann2011knowledge} puts forward a graphical model.

  \textbf{MIML} \citep{surdeanu2012multi} proposes a multi-instance multi-label model.

  \textbf{PCNN} \citep{zeng2015distant} puts forward a piece-wise CNN for relation extraction.

  \textbf{PCNN+ATT} \citep{lin2016neural} proposes the selective attention mechanism with PCNN.

  \textbf{BGRU} \citep{zhou2016attention} proposes a BGRU with the word-level attention mechanism.

  \begin{figure}[htbp]
    \centering
    \includegraphics[width=7.7cm]{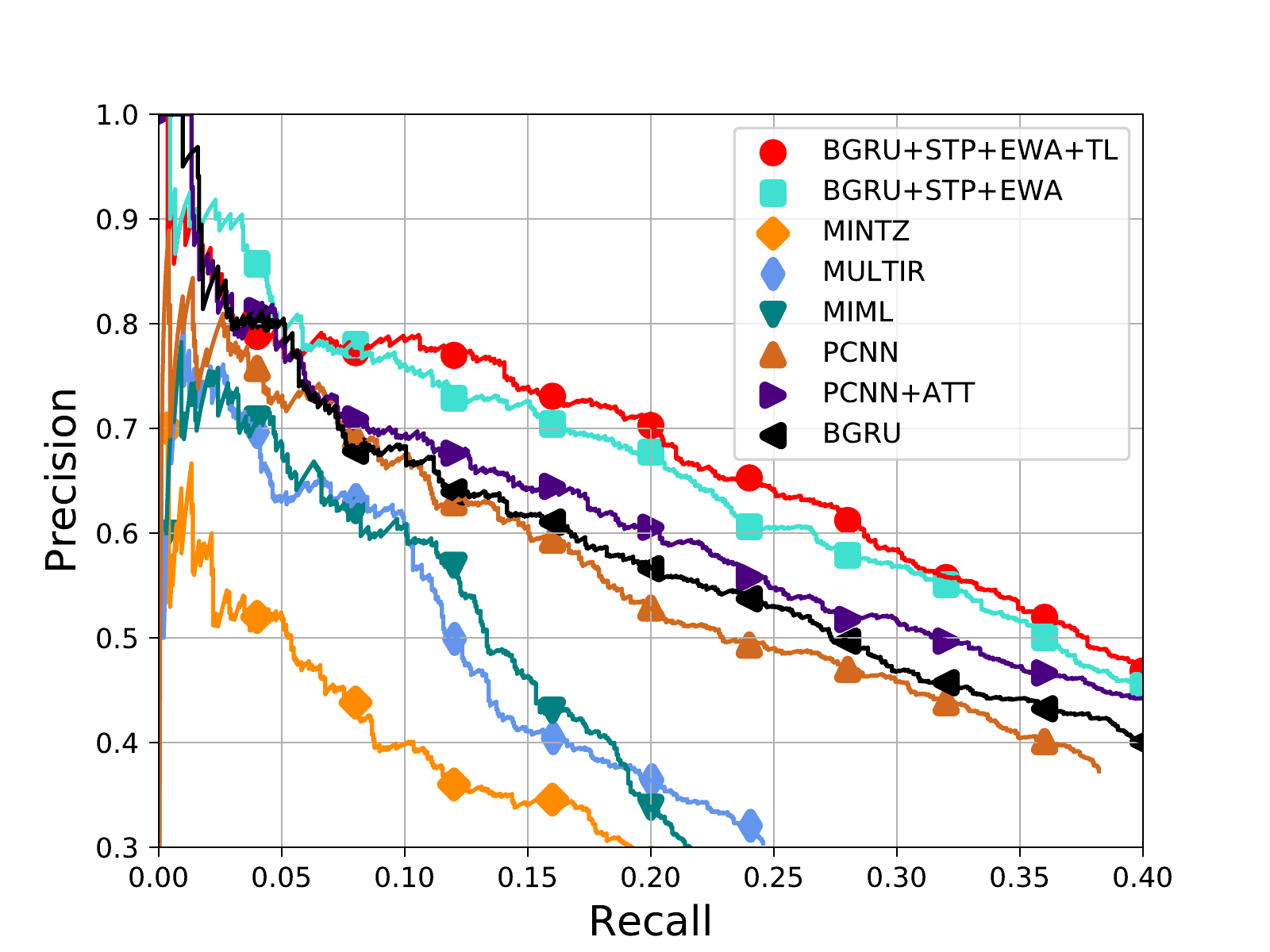}
    \caption{Performance comparison of the proposed method with baselines.}
    \label{fig:model_com}
  \end{figure}

  As Figure~\ref{fig:model_com} shows, we can observe: (1) BGRU+STP+EWA achieves the best PR curve over baselines, which improves the area to $0.38$ over $0.33$ of PCNN, $0.34$ of BGRU and $0.35$ of PCNN+ATT. At the recall rate of $0.25$, our model can still achieve a precision rate above $0.6$. It demonstrates that BGRU+STP+EWA is effective because the STP and entity-wise attention combined with word-level attention can reduce inner-sentence noise at a fine-grained level. (2) Integrated with transfer learning, BGRU+STP+EWA+TL performs much better and increases the PR curve area to $0.392$. It means that the model is pre-trained for better parameter initialization so the TL model becomes more robust against noisy words. Parameter transfer learning can be applied in better feature extractors for further improvement.

  Following previous works, we adopt P@N as a quantitative indicator to compare our model with baselines based on various instances under each relational tuple. In Table~\ref{table:P@N}, we report P@100, P@200, P@300 and the mean of them for each model in the held-out evaluation. We can find: (1) Compared with baselines, BGRU+STP+EWA+TL achieves the best performance in all test settings, which increases the performance of PCNN+ATT in three settings by $6.3\%$, $7.6\%$, and $7.7\%$ respectively. It demonstrates that the integrated model is the most effective; (2) Our STP and entity-wise attention combined with word-level attention reduce inner-sentence noise effectively, and outperform baselines by over $5\%$; (3) Our neural extractor initialized with a priori knowledge learned from entity type classification is more robust against word-level noise where BGRU+STP+EWA+TL has an improvement of $2\%$ over BGRU+STP+EWA.

\section{Conclusion}
  In this paper, we propose a novel word-level approach for distant supervised relation extraction. It aims at tackling the low-quality corpus by reducing inner-sentence noise and improving the robustness against noisy words. To alleviate the influence of word-level noise, we propose the STP. Meanwhile, entity-wise attention combined with word-level attention helps the model focus more on relational words. Furthermore, parameter transfer learning makes our model more robust against noise by reasonable initialization of parameters. The experimental results show that our model significantly and consistently outperforms the state-of-the-art method.

  In the future, we will incorporate the SDP and STP to obtain more precise shortened sentences. Furthermore, we will conduct research in how to utilize entity information to assign more appropriate initial parameters of the relation extractor.

\section*{Acknowledgments}
This work is supported by FDCT 0007/2018/A1, DCT-MoST Joint-project No. (025/2015/AMJ) of SAR Macau; University of Macau Funds Nos: CPG2018-00032-FST \& SRG2018-00111-FST; Chinese National Research Fund (NSFC) Key Project No. 61532013 and National China 973 Project No. 2015CB352401.

\bibliography{emnlp-BTGRU}
\bibliographystyle{acl_natbib_nourl}

\clearpage
%\section*{Acknowledgments}

%\appendix
%\section{Supplemental Material}
%\label{sec:supplemental}

\end{document}